\documentclass{article}
\usepackage{spconf,amsmath,graphicx,hyperref,amssymb,amsfonts}

\usepackage{makecell}
\usepackage{algorithmic}
\usepackage{multirow}
\usepackage{dirtree}
\usepackage{tabularx}
\usepackage{float}
\usepackage{booktabs}
\usepackage{makecell}

\usepackage{url}
\usepackage{graphicx}
\usepackage{textcomp}
\usepackage{xcolor}
\def\BibTeX{{\rm B\kern-.05em{\sc i\kern-.025em b}\kern-.08em
    T\kern-.1667em\lower.7ex\hbox{E}\kern-.125emX}}
\usepackage{booktabs,tabularx,siunitx}
\usepackage{stfloats}

\begin{document}

\title{TRI-DEP: A Trimodal Comparative Study for Depression Detection Using Speech, Text, and EEG}

\name{
    Annisaa Fitri Nurfidausi\sthanks{Both authors contributed equally.}, 
    Eleonora Mancini\textsuperscript{*}, 
    Paolo Torroni
}
\address{DISI, University of Bologna, Italy}

\newcommand{\EMred}[1]{\textcolor{red}{\textbf{EM}: #1}}
\newcommand{\AN}[1]{\textcolor{blue}{\textbf{AN}: #1}}
\ninept
\maketitle

\begin{abstract}
Depression is a widespread mental health disorder, yet its automatic detection remains challenging. Prior work has explored unimodal and multimodal approaches, with multimodal systems showing promise by leveraging complementary signals. However, existing studies are limited in scope, lack systematic comparisons of features, and suffer from inconsistent evaluation protocols. We address these gaps by systematically exploring feature representations and modelling strategies across EEG, together with speech and text. We evaluate handcrafted features versus pre-trained embeddings, assess the effectiveness of different neural encoders, compare unimodal, bimodal, and trimodal configurations, and analyse fusion strategies with attention to the role of EEG. Consistent subject-independent splits are applied to ensure robust, reproducible benchmarking. Our results show that (i) the combination of EEG, speech and text modalities enhances multimodal detection, (ii) pretrained embeddings outperform handcrafted features, and (iii) carefully designed trimodal models achieve state-of-the-art performance. Our work lays the groundwork for future research in multimodal depression detection.
\end{abstract}


\vspace{-0.2cm}
\section{Introduction}
\vspace{-0.2cm}

Depression is a widespread mental health condition predicted to become the second leading cause of disease burden by 2030 \cite{mathers2006projections}, with COVID-19 causing a 27.6 $\%$ rise in global cases \cite{santomauro2021global}. 
In recent years, there has been growing interest in developing automatic depression detection systems to support clinical decision-making and enable telemedicine applications. More recently, multimodal approaches have gained particular attention, motivated by the fact that in clinical settings, such as diagnostic interviews, human expression is inherently multimodal, spanning speech, language, and neural activity. However, current studies often suffer from critical methodological gaps, including limited modality integration, inconsistent evaluation protocols, and potential data leakage, which hinder reproducibility and the fair assessment of model performance.
Models that leverage two modalities dominate the field.
Notable examples include~\cite{yousufi2024multimodal}, who applied DenseNet121 to EEG and speech spectrograms from the MODMA dataset, and~\cite{qayyum2020hybrid}, who employed Vision Transformers on comparable EEG–speech data from MODMA. Other bimodal studies investigated EEG–speech integration with graph convolutional networks~\cite{jia2025multimodal}, speech–text fusion on the E-DAIC dataset using CNN-LSTM attention~\cite{Nykoniuk2025}, and EEG–facial expression fusion~\cite{Tiwary2023}. In~\cite{daly2025depression}, an extensive speech–text comparative analysis with multiple fusion techniques was conducted, but EEG was entirely excluded. Overall, state-of-the-art performances in multimodal depression detection span roughly 85–97\%, depending on the dataset and modality combinations.
All the aforementioned approaches only comprise two modalities, constraining their potential by overlooking trimodal approaches. Moreover, most of them exclude text modality and lack transparent data-splitting protocols.  
In~\cite{he2024dpdnet}, speech, EEG, and text were integrated using GAT-CNN-MpNet architectures on MODMA, achieving about 90$\%$ balanced performance through weighted late fusion, though without comparing handcrafted and pretrained features and with only basic fusion strategies explored. Moreover, the study did not clarify whether 5-fold cross-validation was performed at the segment or subject level.
Our work addresses key limitations in multimodal depression detection by systematically exploring feature representations and modeling strategies across EEG, together with speech and text. We perform a complete comparative analysis of handcrafted features and pretrained embeddings, including, for the first time, brain-pretrained models, evaluate multiple deep learning architectures, and compare unimodal, bimodal, and trimodal configurations. We further investigate how different fusion strategies impact detection accuracy and robustness, with particular attention to the role of EEG. Using consistent subject-independent data splits to ensure reproducible benchmarking, we demonstrate that carefully designed trimodal models achieve state-of-the-art performance. Our study lays the groundwork for the future of multimodal depression detection, guiding the development of more accurate and robust systems. We make both the code and the model checkpoints available to foster transparency and reproducibility.\footnote{\url{https://github.com/annisaafitrinn/tri-dep1}}
\vspace{-0.2cm}

\section{Methodology}
This section describes our experimental approach for trimodal depression detection. We detail the dataset characteristics, preprocessing pipelines for each modality, feature extraction methods encompassing both handcrafted descriptors and pretrained embeddings, modality-specific neural architectures, and multimodal fusion strategies employed throughout our study.
\vspace{-0.2cm}
\subsection{Data}
\vspace{-0.15cm}
This study employs the Multi-modal Open Dataset for Mental-disorder Analysis (MODMA)~\cite{modma}, which provides: (1) 5-minute resting-state EEG recorded with a 128-channel HydroCel Geodesic Sensor Net at 250\,Hz, and (2) audio from structured clinical interviews. For each subject, the interview audio consists of $R=29$ separate recordings (question$-$answer items) whose durations vary across and within subjects (the total interview time is approximately 25 minutes per subject). Since MODMA does not include text transcriptions from clinical interviews, we generate automatic transcriptions using speech-to-text models. The dataset comprises individuals diagnosed with Major Depressive Disorder (MDD), recruited from Lanzhou University Second Hospital, and Healthy Controls (HC) obtained via public advertising; MDD diagnoses were confirmed by licensed psychiatrists. In this study, we retain only subjects who participated in both EEG and interview recordings, resulting in a filtered cohort of 38 subjects. Table~\ref{tab:participant_demographics} summarises demographic information across groups and protocols. Additional details are available in~\cite{modma}.

\begin{table}[htbp]
\centering
\caption{Participant demographics in the MODMA dataset. 
\textit{M}: Male, \textit{F}: Female, \textit{HC}: Healthy Control, and \textit{MDD}: Major Depressive Disorder.}
\resizebox{0.9\columnwidth}{!}{%
\begin{tabular}{l c c c c}
\toprule
\textbf{Modality} & \textbf{Total} & \textbf{MDD} & \textbf{HC} & \textbf{Age} \\
                  &                & \textbf{(M/F)} & \textbf{(M/F)} & \textbf{MDD/HC)} \\
\midrule
128-ch EEG & 53 & 24 (13/11) & 29 (20/9) & 16--56 / 18--55 \\
Speech     & 52 & 23 (16/7)  & 29 (20/9) & 16--56 / 18--55 \\
\bottomrule
\end{tabular}%
}
\label{tab:participant_demographics}
\end{table}

MODMA's compliance with ethical guidelines is detailed by~\cite{mancini2024promoting}, demonstrating satisfaction of core criteria (informed consent, ethical approval, clinical validation). While certain desiderata (e.g., educational level, ethnicity) are not fully met, essential requirements are satisfied.

Many studies lack clarity in data splitting \cite{qayyum2023high, yousufi2024multimodal, he2024dpdnet}, where segment-level splits can leak information by placing recordings from the same subject in both training and test sets, yielding inflated performance. To avoid this, we use stratified 5-fold subject-level cross-validation with consistent splits across experiments. We also release these splits on our companion website\footnote{\url{https://annisaafitrinn.github.io/tri-dep/}} to ensure reproducibility and fair comparison. 

To address the lack of transcriptions in the MODMA dataset, we employ WhisperX~\cite{bain23_interspeech} to generate text for each subject's 29 recordings, without further post-processing. We acknowledge that Automatic Speech Recognition (ASR) systems are known to exhibit reduced accuracy on clinical speech~\cite{Miner2020,Just2025}, potentially introducing transcription errors; however, ground-truth transcriptions are not available for verification.

\subsection{Experimental Pipeline Design}
We design a unified pipeline for multimodal depression detection with EEG, speech, and text.
For EEG, we adopt two processing branches: a 29-channel, 250 Hz, 10 s segmentation setup, consistent with prior work~\cite{qayyum2023high, yousufi2024multimodal, he2024dpdnet, khan2024machine}, and a 19-channel, 200 Hz, 5 s segmentation setup replicating the preprocessing used in CBraMod~\cite{wang2024cbramod} for the MUMTAZ depression dataset~\cite{mumtaz2019deep}
. For CBraMod, we evaluated both the original pre-trained version and the model fine-tuned on MUMTAZ, as described in the official documentation\footnote{\scriptsize{\url{https://github.com/wjq-learning/CBraMod/blob/main/}}}
, and found the latter consistently superior. Therefore, throughout this work we refer to CBraMod as the MUMTAZ-fine-tuned model.
Speech recordings are resampled to 16 kHz, denoised, and segmented into 5 s windows with 50\% overlap, while text is used directly from raw Chinese transcriptions.

Feature extraction combines handcrafted descriptors (EEG statistics, spectral power, entropy; speech MFCCs with/without prosody) with embeddings from large pre-trained models. For EEG, we employ both the Large Brain Model (LaBraM)~\cite{jiang2024labram}, trained on $\sim$2,500 hours of EEG from 20 datasets, and CBraMod, a patch-based masked reconstruction model. For speech, we use XLSR-53~\cite{conneau2020xlsr}, a multilingual wav2vec~2.0 encoder, and Chinese HuBERT Large~\cite{tencentgamemate_chinese_hubert_large}, trained on 10k hours of WenetSpeech. For text, we use Chinese BERT Base~\cite{devlin2019bert}, MacBERT~\cite{cui-etal-2020-revisiting}, XLNet~\cite{yang2019xlnet}, and MPNet Multilingual~\cite{reimers-2019-sentence-bert}. Segment-level representations are encoded with a combination of CNNs, LSTMs, and/or GRUs (with/without attention) and fused using decision-level strategies.

\vspace{-0.15cm}

\subsection{Data Preprocessing}
The preprocessing stage serves multiple objectives, including cleaning and structuring the raw data, as well as preparing it for multimodal analysis. 
One key objective is the segmentation of the input into smaller units that can be more effectively processed by the models. 
We denote with $\mathbf{S}_{\text{EEG}}$, $\mathbf{S}_{\text{SPEECH}}$, and $\mathbf{S}_{\text{TEXT}}$ the number of segments obtained after preprocessing for each input modality.

\vspace{2mm}
\noindent{\textbf{EEG ---}} 
For handcrafted features and LaBraM, we follow prior work~\cite{qayyum2023high, yousufi2024multimodal, he2024dpdnet, khan2024machine}, which comprises retaining $C=29$ channels\footnote{\label{fn:channels}Full list available on our companion website.}, applying a 0.5$-$50\,Hz bandpass filter with a 50\,Hz notch, and average re-referencing. Recordings are segmented into 10\,s windows; at 250\,Hz, each window contains $T = 250 \times 10 = 2500$ samples. Thus, a recording of length $L$ seconds produces $S_{\text{EEG}} = L/10$ windows (e.g., $S_{\text{EEG}}=30$ for a 5-min recording), represented as $\mathbf{X}_{\text{EEG}}^{(1)} \in \mathbb{R}^{S_{\text{EEG}} \times C \times T}$.  

For CBraMod, we use the version pretrained on the MUMTAZ depression dataset, thereby replicating its preprocessing. Signals are resampled to 200\,Hz, bandpass filtered (0.3$-$75\,Hz) with a 50\,Hz notch, and reduced to $C=19$ channels\footnotemark[\getrefnumber{fn:channels}]. Recordings are segmented into 5\,s windows; at 200\,Hz, each window contains $T = 200 \times 5 = 1000$ samples. A recording of length $L$ seconds thus yields $S_{\text{EEG}} = L/5$ windows (e.g., $S_{\text{EEG}}=60$ for a 5-min recording). Each window is further divided into $P=5$ non-overlapping patches of $T_\text{patch}=200$ samples, resulting in $\mathbf{X}_{\text{EEG}}^{(2)} \in \mathbb{R}^{S_{\text{EEG}} \times C \times P \times T_\text{patch}}$. 

\vspace{2mm}
\noindent{\textbf{Speech ---}} Audio recordings are resampled from 44\,kHz to 16\,kHz, converted to mono PCM, amplitude-normalized to $[-1,1]$, silence-trimmed, and denoised with a median filter~\cite{gheorghe2023using}. Each signal is segmented into overlapping windows of length $w=5$\,s with hop size $h=2.5$\,s (50\% overlap). At a sampling rate of 16\,kHz, each segment contains $T_{\text{seg}}=80{,}000$ samples and each hop $T_{\text{hop}}=40{,}000$ samples.

For a recording of duration $L$ seconds (post-trimming), the number of segments is 
$S_{\text{SPEECH}}=\big\lfloor (L-w)/h \big\rfloor + 1$ for $L \ge w$, while recordings shorter than $w$ are retained as a single segment. The segmented waveform is represented as $\mathbf{X}_{\text{SPEECH}} \in \mathbb{R}^{S_{\text{SPEECH}} \times T_{\text{seg}}}$, where each row corresponds to one waveform segment. Each subject has $R=29$ interview recordings; after windowing, recording $r$ yields $S_{\text{SPEECH}}^{(r)}$ segments $\mathbf{X}_{\text{SPEECH}}^{(r)} \in \mathbb{R}^{S_{\text{SPEECH}}^{(r)} \times T_{\text{seg}}}$. The subject-level speech representation is the concatenation along the segment axis:
$\mathbf{X}_{\text{SPEECH}}=\big[\mathbf{X}_{\text{SPEECH}}^{(1)};\dots;\mathbf{X}_{\text{SPEECH}}^{(R)}\big]$,
with a total of $S_{\text{SPEECH}}=\sum_{r=1}^{R} S_{\text{SPEECH}}^{(r)}$ segments.

\vspace{2mm}
\noindent{\textbf{Text ---}} 
Each recording has a single transcript. After tokenisation, the subject-level text representation is the concatenation of all transcript representations,
$\mathbf{X}_{\text{TEXT}}=\big[\mathbf{X}_{\text{TEXT}}^{(1)};\dots;\mathbf{X}_{\text{TEXT}}^{(R)}\big]$. 

\subsection{Feature Extraction}
\noindent{\textbf{EEG ---}} \emph{Handcrafted features.}  
For each segment $\mathbf{X}_{\text{EEG}}^{(1)} \in \mathbb{R}^{C \times T}$ we extract $F=10$ handcrafted descriptors per channel 
(statistical, spectral, entropy), yielding 
$\mathbf{X}_{\text{HAND}} \in \mathbb{R}^{S \times C \times F}$.  

\emph{Pre-trained models.}  
We further extract embeddings from LaBraM and 
CBraMod. 
LaBraM operates on $\mathbf{X}_{\text{EEG}}^{(1)}$ and maps each segment to a 
$D=200$-dimensional embedding, producing 
$\mathbf{X}_{\text{LaBraM}} \in \mathbb{R}^{S \times D}$.  
CBraMod operates on $\mathbf{X}_{\text{EEG}}^{(2)}$, where each 5\,s segment is 
patch-encoded and then averaged across channels and patches to form 
$D=200$-dimensional embeddings, resulting in 
$\mathbf{X}_{\text{CBraMod}} \in \mathbb{R}^{S \times D}$.  

\emph{Remark.}  
After feature extraction, the raw temporal dimension ($T$ or $T_{\text{patch}}$) 
is no longer present, as each segment is reduced to a fixed-size representation 
of dimension $F$ (handcrafted) or $D$ (embeddings).
For subject-level modelling, features from all recordings of the same subject 
are stacked to form the final subject representation.
\vspace{2mm} 

\noindent{\textbf{Speech ---}} From each waveform segment, we compute two handcrafted variants, namely MFCCs (40 coefficients) and Prosody $+$ MFCCs (46 features: 40 MFCCs plus energy, $F_0$, RMS energy, pause rate, phonation time, speech rate). We also extract segment embeddings with XLSR-53 and Chinese HuBERT Large. 
Segment-level features are stacked per recording and then concatenated across the 29 recordings of each subject to form the subject-level representation.
The exact tensor shapes (per recording and subject-level) are summarised in Table~\ref{tab:speech_feature_shapes}. 
After feature extraction, the raw sample length is no longer present; each segment is represented by a fixed-size vector (40/46/768/1024 dimensions).

\begin{table}[!ht]
\centering
\caption{Shapes of speech feature matrices per recording $r$ and at the subject level.}
\label{tab:speech_feature_shapes}
\small
\vspace{1mm}
\begin{tabular}{@{}lcc@{}}
\toprule
\textbf{Feature name} & \textbf{Per-recording shape} & \textbf{Subject-level shape} \\
\midrule
$\mathbf{X}_{\text{MFCCs}}$ & $\mathbb{R}^{S_{\text{SPEECH}}^{(r)} \times 40}$ & $\mathbb{R}^{S_{\text{SPEECH}} \times 40}$ \\
$\mathbf{X}_{\text{PROSODY+MFCCs}}$ & $\mathbb{R}^{S_{\text{SPEECH}}^{(r)} \times 46}$ & $\mathbb{R}^{S_{\text{SPEECH}} \times 46}$ \\
$\mathbf{X}_{\text{XLSR}}$ & $\mathbb{R}^{S_{\text{SPEECH}}^{(r)} \times 1024}$ & $\mathbb{R}^{S_{\text{SPEECH}} \times 1024}$ \\
$\mathbf{X}_{\text{HuBERT}}$ & $\mathbb{R}^{S_{\text{SPEECH}}^{(r)} \times 768}$ & $\mathbb{R}^{S_{\text{SPEECH}} \times 768}$ \\
\bottomrule
\end{tabular}
\end{table}

\noindent{\textbf{Text ---}}
\label{sec:text_features}
Each recording has a single transcript, which we encode with a pretrained language model (BERT, MacBERT, XLNet, or MPNet) to obtain one \(D=768\)-dimensional embedding per recording; for a subject with \(R=29\) recordings, stacking these yields \(\mathbf{X}_{\text{BERT}}, \mathbf{X}_{\text{MacBERT}}, \mathbf{X}_{\text{XLNet}}, \mathbf{X}_{\text{MPNet}} \in \mathbb{R}^{R \times D}\) (with \(R=29\), \(D=768\)).

\subsection{Baselines}
We re-implement two multimodal baselines for depression detection that use standard image architectures on EEG and speech \emph{2D-spectrograms (Spec2D)}: DenseNet-121~\cite{yousufi2024multimodal} and Vision Transformer (ViT)~\cite{qayyum2020hybrid}. These studies are among the few that explore EEG$-$speech multimodality in this task and report promising results. In our experiments, we retain their model architectures but apply our own subject-level cross-validation splits for consistency, making results not directly comparable to the original works. Additional implementation details are provided on our companion website.

\subsection{Architectures}
We assess several modality-tailored architectures.  We also experiment with multimodality, combining predictions from the best-performing feature-model pair in each modality in a late fusion fashion.

To keep notation light, we use \(\mathbf{F}\) to denote the \emph{generic feature matrix} per modality, either handcrafted features or embeddings from pretrained models.
Concretely, \(\mathbf{F}_{\text{EEG}}\) (EEG), \(\mathbf{F}_{\text{SPEECH}}^{(r)}\) (speech, per recording \(r\)), and \(\mathbf{F}_{\text{TEXT}}\) (text, subject-level).

\vspace{2mm}
\noindent\textbf{EEG ---}
We consider three detection modules: \textit{CNN+LSTM}, \textit{GRU+Attention}, and \textit{CNN}.
The CNN+LSTM branch uses two 1D convolutions (kernel size 3, padding 1) with dropout to capture local temporal patterns, followed by a 2-layer LSTM for sequence modelling.
\textit{GRU+Attention} uses a 2-layer GRU with an attention mechanism that weights hidden states to form a subject-level summary.
\textit{CNN} retains only the convolutional layers followed by max-pooling and an MLP head, dispensing with recurrence entirely.
All three modules consume \(\mathbf{F}_{\text{EEG}}\) and produce a latent representation \(\mathbf{H}_{\text{EEG}}\); an MLP head outputs \(y_{\text{EEG}}\).

\vspace{2mm}
\noindent\textbf{Speech ---}
A shallow CNN extracts segment-level features from each recording. 
These are reduced to a single fixed-size vector 
\(\mathbf{F}_{\text{SPEECH}}^{(r)} \in \mathbb{R}^{d}\) 
using one of three encoders: 
(i) max pooling, 
(ii) GRU with attention, or 
(iii) BiGRU with attention (the latter extending the GRU+Attn design with bidirectional recurrence). 
The resulting \(R=29\) vectors are stacked into the subject-level matrix 
\(\mathbf{F}_{\text{SPEECH}} \in \mathbb{R}^{R \times d}\). 
This sequence is then processed by a subject-level classifier---either an LSTM or a CNN with max-pooling---to produce the subject-level representation 
\(\mathbf{H}_{\text{SPEECH}}\), 
which is fed to an MLP head to obtain the final prediction 
\(y_{\text{SPEECH}}\).

\vspace{2mm}

\noindent\textbf{Text ---}
\(\mathbf{F}_{\text{TEXT}}\) denotes the subject-level text features (Sec.~\ref{sec:text_features}).
A detection module (LSTM or CNN) transforms \(\mathbf{F}_{\text{TEXT}}\) into \(\mathbf{H}_{\text{TEXT}}\), and an MLP head outputs \(y_{\text{TEXT}}\).

\vspace{2mm}

\noindent\textbf{Multimodal Fusion ---} We select, for each modality, the best-performing feature-model pair and fuse their predictions via decision-level (late) fusion. This design choice is motivated by modularity and interpretability: each modality is trained independently with its own optimal architecture, and fusion operates on output predictions, making it straightforward to add or remove modalities without retraining. This property is particularly valuable in clinical settings, where not all modalities may be available for every patient. We consider three schemes: \textit{weighted averaging} -- compute weighted combination of modality probabilities where weights sum to one, then predict the class with highest weighted probability; \textit{Bayesian fusion} -- convert modality-specific posteriors to likelihood ratios, combine them with predefined weights, and map back to a posterior; \textit{majority voting} -- each modality independently predicts a class label, and the final prediction is the class receiving the most votes, with ties resolved arbitrarily at 0.5.
\section{Experimental Setup}
We adopt stratified 5-fold cross-validation with fixed subject splits to ensure balanced and comparable experiments and prevent data leakage. Models are trained with cross-entropy loss and softmax output, with hyperparameters tuned manually. All implementation details are provided on our companion website.
Given the limited sample size (38 subjects, 3--4 validation subjects per fold), early stopping criteria based on validation metrics are unreliable, since a single misclassification shifts validation accuracy by 25--33\%. To avoid overfitting to noisy validation signals, all models are trained for a fixed number of epochs, determined per configuration through preliminary experiments on the validation set. At the end of training, the final model state is used for inference. This ensures a consistent and reproducible training protocol across all modalities and configurations.

\section{Results}
In this section, we report the performance of all experimental categories: baseline re-implementations, unimodal models, and our proposed multimodal architectures. F1-scores are reported as mean $\pm$ standard deviation across folds. Table~\ref{tab:unimodal_results} presents the baselines and unimodal models, including the best-performing model for each set of features per modality. Table~\ref{tab:multimodal_results} reports the performance of baseline models and multimodal fusion strategies, highlighting the best configuration within each category and the overall best-performing model. Further results are available on our companion website.

\begin{table*}[tp]
\centering
\scriptsize
\renewcommand{\arraystretch}{1.15}
\setlength{\tabcolsep}{0.5pt}
\caption{Results of baselines and unimodal models (macro-F1, mean $\pm$ std across 5 folds). 
The \textit{Features} column indicates the input representation (handcrafted or learned); the \textit{Detection Module} column indicates the classifier trained on top. For speech, the detection module comprises a segment-level encoder and a subject-level classifier, separated by $\rightarrow$.
In \textbf{bold}, the best performing configuration per modality.}
\vspace{1mm}
\resizebox{0.9\linewidth}{!}{%
\begin{tabular}{
  >{\centering\arraybackslash}m{2.5cm}
  >{\raggedright\arraybackslash}m{2.5cm}
  >{\centering\arraybackslash}m{7.0cm}
  >{\centering\arraybackslash}m{2.5cm}
}
\toprule
\textbf{Category} & \textbf{Features} & \textbf{Detection Module} & \textbf{F1} \\
\midrule
\multirowcell{2}[0pt][c]{Baselines \\ (Speech+EEG)}
  & $\mathbf{X}_{\text{Spec2D}}$   & ViT           &  0.560 $\pm$ 0.190 \\
  & $\mathbf{X}_{\text{Spec2D}}$   & DenseNet-121  & 0.586 $\pm$ 0.240 \\
\midrule
\multirowcell{3}[0pt][c]{EEG}
  & $\mathbf{X}_{\text{HAND}}$     & CNN+LSTM      & 0.418 $\pm$ 0.096 \\
  & $\mathbf{X}_{\text{LaBraM}}$   & GRU+Attn      & 0.417 $\pm$ 0.089 \\
  & $\mathbf{X}_{\text{CBraMod}}$  & \textbf{CNN}  & $\mathbf{0.446 \pm 0.106}$ \\
\midrule
\multirowcell{4}[0pt][c]{Speech}
  & $\mathbf{X}_{\text{MFCCs}}$           & CNN+MaxPool $\rightarrow$ LSTM                     & 0.553 $\pm$ 0.143 \\
  & $\mathbf{X}_{\text{Prosody+MFCCs}}$   & CNN+BiGRU+Attn $\rightarrow$ LSTM                  & 0.655 $\pm$ 0.182 \\
  & $\mathbf{X}_{\text{XLSR}}$           & CNN+MaxPool $\rightarrow$ CNN+MaxPool               & 0.774 $\pm$ 0.143 \\
  & $\mathbf{X}_{\text{HuBERT}}$         & \textbf{CNN+BiGRU+Attn $\rightarrow$ CNN+MaxPool}  & $\mathbf{0.809 \pm 0.081}$ \\
\midrule
\multirowcell{4}[0pt][c]{Text}
  & $\mathbf{X}_{\text{MPNet}}$   & CNN   & 0.725 $\pm$ 0.244 \\
  & $\mathbf{X}_{\text{BERT}}$    & CNN   & 0.720 $\pm$ 0.122 \\
  & $\mathbf{X}_{\text{XLNet}}$   & LSTM  & 0.665 $\pm$ 0.088 \\
  & $\mathbf{X}_{\text{MacBERT}}$ & \textbf{LSTM} & \textbf{0.784 $\pm$ 0.114} \\
\bottomrule
\end{tabular}%
}
\label{tab:unimodal_results}
\end{table*}

\begin{table*}[tp]
\centering
\scriptsize
\renewcommand{\arraystretch}{1.15}
\setlength{\tabcolsep}{1pt}
\caption{Baseline and multimodal fusion results (F1-score, mean $\pm$ std across 5 folds). 
In \textbf{bold}, the best performing configuration per category. 
The overall best are additionally \underline{underlined}. 
Fusion weights were optimised via grid search (step\,=\,0.05).}
\vspace{1mm}
\resizebox{0.8\linewidth}{!}{%
\begin{tabular}{
  >{\centering\arraybackslash}m{3.0cm}
  >{\raggedright\arraybackslash}m{6.4cm}
  >{\centering\arraybackslash}m{3.5cm}
}
\toprule
\textbf{Category} & \textbf{Configuration} & \textbf{F1-score} \\
\midrule
\multirowcell{2}[0pt][c]{Baselines}
  & ViT & 0.560 $\pm$ 0.190 \\
  & \textbf{DenseNet-121}  & $\mathbf{0.586 \pm 0.240}$
 \\
\midrule
\multirowcell{4}[0pt][c]{Weighted Averaging}
  & EEG + Speech (0.05 : 0.95) & 0.809 $\pm$ 0.081 \\
  & EEG + Text (0.15 : 0.85) & 0.786 $\pm$ 0.116 \\
  & Speech + Text (0.45 : 0.55) & 0.839 $\pm$ 0.059 \\
  & \textbf{\underline{EEG + Speech + Text (0.05 : 0.35 : 0.60)}} & \textbf{\underline{0.864 $\pm$ 0.095}} \\
\midrule
\multirowcell{4}[0pt][c]{Bayesian Fusion}
  & EEG + Speech (0.05 : 0.95) & 0.809 $\pm$ 0.081 \\
  & EEG + Text (0.05 : 0.95) & 0.784 $\pm$ 0.114 \\
  & Speech + Text (0.20 : 0.80) & 0.863 $\pm$ 0.140 \\
  & \textbf{\underline{EEG + Speech + Text (0.05 : 0.35 : 0.65)}} & \textbf{\underline{0.864 $\pm$ 0.095}} \\
\midrule
\multirowcell{4}[0pt][c]{Majority Voting}
  & EEG + Speech & 0.557 $\pm$ 0.135 \\
  & EEG + Text & 0.500 $\pm$ 0.078 \\
  & \textbf{Speech + Text} & \textbf{0.809 $\pm$ 0.081} \\
  & EEG + Speech + Text & 0.800 $\pm$ 0.200 \\
\bottomrule
\end{tabular}%
}

\label{tab:multimodal_results}
\end{table*}

\vspace{1mm}\noindent\textbf{Unimodal ---} Table~\ref{tab:unimodal_results} reports the performance of baseline and unimodal models. Among EEG features, CBraMod embeddings with a CNN classifier achieve the best result (F1\,=\,0.446), confirming the benefit of fine-tuning on a depression-related corpus, though all EEG configurations remained below 0.5 in terms of macro F1-score. For speech, Chinese HuBERT embeddings coupled with a BiGRU encoder and CNN+MaxPool detection module achieve the strongest performance (F1\,=\,0.809), while handcrafted MFCCs and prosodic features yield lower scores, indicating that deep speech embeddings capture richer information. In the text modality, Chinese MacBERT with an LSTM reach the top result (F1\,=\,0.784). Overall, speech provides the most informative single modality, closely followed by text, while EEG remained less predictive in isolation.

\vspace{1mm}\noindent\textbf{Multimodal ---} Table~\ref{tab:multimodal_results} compares the baselines with different fusion strategies. Simple baselines such as ViT and DenseNet-121 reach F1-scores around 0.56. Fusion weights are optimised via grid search (step\,=\,0.05) over all modality combinations. The trimodal configurations achieve the highest macro F1-score for both weighted averaging (0.864) and Bayesian fusion (0.864), outperforming the best unimodal model (speech, F1\,=\,0.809) by approximately 5.5 percentage points. Among bimodal configurations, Speech+Text consistently perform best across all fusion methods, with weighted averaging achieving F1\,=\,0.839 and notably low variance (std\,=\,0.059). These results confirm the complementarity of modalities: while speech and text individually provide strong unimodal performance, integrating all three modalities yields the strongest overall result.

\begin{figure}[H]
  \centering
  \includegraphics[width=1.0\columnwidth]{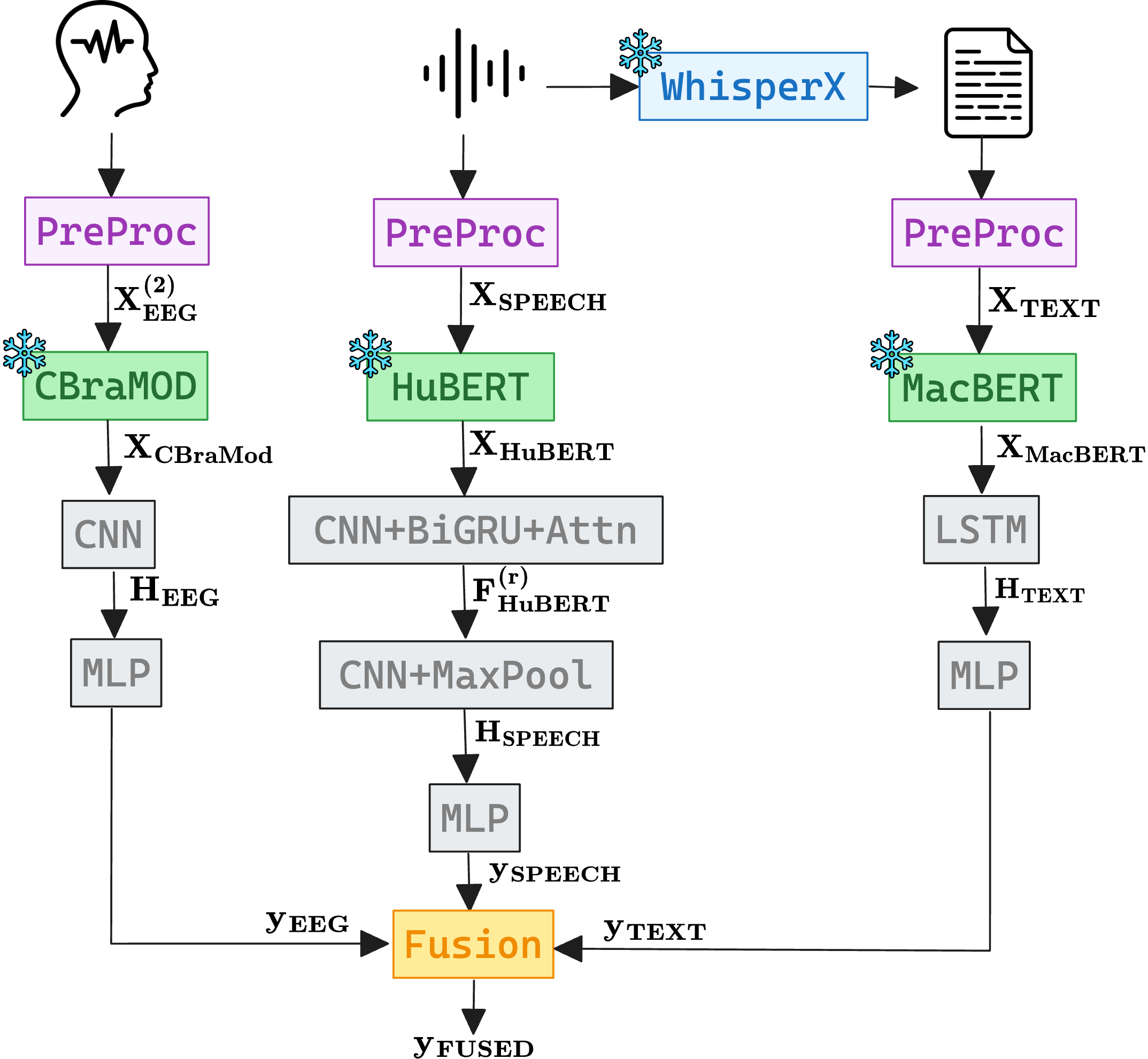}
  \caption{Experimental Framework for Multimodal Depression Detection}
  \label{fig:architecture}
\end{figure}

\vspace{2mm}\noindent\textbf{Experimental Framework for Multimodal Depression Detection ---} Building on this systematic exploration of feature extraction methods, neural architectures, and fusion strategies, we propose an experimental framework for multimodal depression detection, illustrated in Figure~\ref{fig:architecture}. 

The framework selects the best-performing predictors for each modality: $\mathbf{X}_{\text{CBraMod}}$ processed with a CNN for EEG, $\mathbf{X}_{\text{HuBERT}}$ processed with a CNN+BiGRU+Attn encoder and CNN+MaxPool classifier for speech, and $\mathbf{X}_{\text{MacBERT}}$ processed with an LSTM for text. These modality-specific pipelines are then combined through alternative decision-level fusion strategies. Our best-performing configurations use either weighted averaging or Bayesian fusion across the three modalities with optimized weights, both achieving a macro-F1 of 0.864 averaged across 5 folds.


We employ two complementary non-parametric tests, following~\cite{dietterich1998approximate}: McNemar's exact test, which evaluates whether two classifiers make significantly different errors on the same subjects ($N{=}38$), and the Wilcoxon signed-rank test, which assesses whether per-fold macro-F1 differences are consistent across folds ($n{=}5$) without assuming normality.

Table~\ref{tab:significance} reports results for 12 selected comparisons
that address three key questions: whether each trimodal fusion method outperforms the best unimodal and best bimodal configurations, whether the best bimodal (weighted averaging fusion between speech and text) outperforms individual modalities, and how the three modalities rank against each other. 
``Best'' refers to the best-performing unimodal configuration per modality 
(bold rows in Table~\ref{tab:unimodal_results}): Best 
Text\,=\,$\mathbf{X}_{\text{MacBERT}}$+LSTM, Best Speech\,=\,$\mathbf{X}_{\text{HuBERT}}$+BiGRU+\allowbreak CNN+\allowbreak MaxPool, Best 
EEG\,=\,$\mathbf{X}_{\text{CBraMod}}$+CNN.

With only $N{=}38$ subjects and $n{=}5$ folds, no comparison reaches statistical 
significance at $\alpha{=}0.05$; notably, the minimum achievable $p$-value for the Wilcoxon test with 5 paired observations is $0.0625$.
Furthermore, although fusion does not yield statistically significant improvements over the best unimodal model at this sample size, it consistently increases macro-F1 and improves stability: weighted averaging Speech+Text achieves
a standard deviation of 0.059, lower than both text alone (0.114) and the trimodal configuration (0.095), indicating more stable predictions across different patient subgroups.
\begin{table}[ht]
\centering
\footnotesize
\renewcommand{\arraystretch}{1.17}
\setlength{\tabcolsep}{3pt}
\caption[Pairwise statistical significance tests for 12 selected model comparisons.]{
Pairwise statistical significance tests for 12 selected comparisons from the 253 total 
pairs across the 23 configurations reported in 
Tables~\ref{tab:unimodal_results}--\ref{tab:multimodal_results}.
$\Delta$F1 is the difference in mean macro-F1 between Model~A and Model~B 
(positive values indicate Model~A outperforms Model~B). 
No comparison reaches significance at $\alpha{=}0.05$.
WA\,=\,Weighted Averaging, BF\,=\,Bayesian Fusion, MV\,=\,Majority Voting,
E\,=\,EEG, Sp\,=\,Speech, Tx\,=\,Text.}
\vspace{1mm}
\begin{tabular}{llccc}
\toprule
\textbf{Model A} & \textbf{Model B} & \textbf{$\Delta$F1} & \textbf{McNemar $p$} & \textbf{Wilcoxon $p$} \\
\midrule
  WA: E+Sp+Tx & Best Text    & +0.081 & 0.6476 & 0.2379 \\
  BF: E+Sp+Tx & Best Text    & +0.081 & 0.7257 & 0.2379 \\
  MV: E+Sp+Tx & Best Text    & +0.016 & 1.0000 & 1.0000 \\
  WA: E+Sp+Tx & Best Speech  & +0.055 & 0.8765 & 0.6286 \\
  BF: E+Sp+Tx & Best Speech  & +0.055 & 0.7673 & 0.6286 \\
  WA: Sp+Tx   & Best Text    & +0.055 & 0.9297 & 0.3906 \\
  WA: Sp+Tx   & Best Speech  & +0.030 & 1.0000 & 1.0000 \\
  WA: E+Sp+Tx & WA: Sp+Tx    & +0.025 & 1.0000 & 1.0000 \\
  BF: E+Sp+Tx & WA: Sp+Tx    & +0.025 & 1.0000 & 1.0000 \\
  Best Text   & Best Speech  & $-$0.025 & 1.0000 & 0.6286 \\
  Best Text   & Best EEG     & +0.337 & 0.1472 & 0.2379 \\
  Best Speech & Best EEG     & +0.363 & 0.1002 & 0.1832 \\
\bottomrule
\end{tabular}

\label{tab:significance}
\end{table}

The framework serves as a reference setup for future experiments, enabling systematic evaluation of new fusion strategies or additional modalities. The limited statistical power observed in our significance analysis is consistent with the broader landscape of clinical affective computing, where datasets typically comprise 30--50 subjects and pairwise significance is rarely achievable. Similar power limitations have been reported in related multimodal depression detection studies, and remain an open challenge for the field. Despite this constraint, the consistent direction of improvement across multiple fusion methods and modality combinations provides converging evidence for the benefit of multimodal integration. To the best of our knowledge, our trimodal configuration with weighted averaging fusion represents state-of-the-art performance in multimodal depression detection.

\section{Conclusion}
We addressed key limitations in multimodal depression detection by adopting subject-level 
stratified cross-validation and systematically exploring feature representations and 
fusion strategies across EEG, speech, and text. Pretrained embeddings consistently 
outperformed handcrafted features across all three modalities. In the unimodal setting, 
speech emerged as the most predictive modality, closely followed by text, while EEG 
remained the least discriminative in isolation. In the multimodal setting, decision-level 
fusion consistently improved over unimodal models, with trimodal configurations achieving 
the strongest results (macro-F1\,=\,0.864). A notable finding is the asymmetric role of 
EEG: despite weak unimodal performance, it provides complementary information that 
benefits trimodal fusion in the presence of text. Bimodal speech+text fusion offers a 
particularly favourable trade-off between performance and stability 
(F1\,=\,0.839, std\,=\,0.059). No pairwise comparison reached significance at 
$\alpha{=}0.05$, consistent with the power limitations of small clinical cohorts, 
though the consistent direction of improvement across fusion methods provides converging 
evidence for multimodal integration. Beyond the best-performing configuration, we 
introduce an experimental framework as a reference setup for future work, enabling 
systematic evaluation of new fusion strategies or additional modalities. A current 
limitation is the reliance on automated transcriptions via WhisperX rather than 
ground-truth text; future work could address this by investigating ASR systems trained 
on clinical speech. All data splits, preprocessing scripts, and code are released to 
support reproducibility and further advances in multimodal depression detection.
This work establishes a rigorous and reproducible benchmark that future research 
can build upon, whether by incorporating additional modalities, investigating 
early or intermediate fusion strategies, improving transcription quality for 
clinical speech, or extending the framework to larger and more diverse cohorts.
\section{Acknowledgments}
This work uses data from the MODMA dataset~\cite{modma} provided by the Gansu Provincial Key Laboratory of Wearable Computing, Lanzhou University, China. We gratefully acknowledge the data contributors for granting access to the data. 

\section{Compliance with Ethical Standards}
This study was conducted retrospectively using human subject data from the MODMA dataset ~\cite{modma}. Access to the dataset is granted by the data owners upon request, and we do not redistribute any data. According to the terms of use specified by the dataset providers, separate ethical approval was not required for our analyses. All experiments were carried out in compliance with these conditions.

\bibliographystyle{IEEEbib}
\bibliography{refs}

\end{document}